\RequirePackage{fix-cm}
\documentclass[twocolumn,final,natbib]{svjour3}
\usepackage{float,graphicx,amssymb,subfig,epsfig,amsmath}
\usepackage[export]{adjustbox}
\usepackage{breakcites}
\usepackage[breaklinks=true]{hyperref}
\newcommand{\figwidth}{3.1in}
\graphicspath{{epsfigs/}}

\journalname{Soft Comput}
\begin{document}

\title{Discrete and fuzzy dynamical genetic programming in the XCSF learning classifier system}

\author{Richard J. Preen \and Larry Bull}

\institute{Richard J. Preen \and Larry Bull \at Department of Computer Science and Creative Technologies \\ University of the West of England, Bristol, BS16 1QY, UK \\ \email{richard2.preen@uwe.ac.uk, larry.bull@uwe.ac.uk}}

\date{The final version of this paper is published in Soft Computing 2014, 18(1):153--167.}

\maketitle

\begin{abstract}
A number of representation schemes have been presented for use within learning classifier systems, ranging from binary encodings to neural networks. This paper presents results from an investigation into using discrete and fuzzy dynamical system representations within the XCSF learning classifier system. In particular, asynchronous random Boolean networks are used to represent the traditional condition-action production system rules in the discrete case and asynchronous fuzzy logic networks in the continuous-valued case. It is shown possible to use self-adaptive, open-ended evolution to design an ensemble of such dynamical systems within XCSF to solve a number of well-known test problems.
\end{abstract}

\section{Introduction}
\label{intro}
\begin{sloppypar}
Traditionally, learning classifier systems \citep[LCS;][]{Holland:1976} use a ternary encoding to generalize over the environmental inputs and to associate appropriate actions. A number of representations have previously been presented beyond this scheme however, including real numbers \citep{Wilson:2000}, fuzzy logic \citep{Valenzuela-Rendon:1991} and artificial neural networks \citep{Bull:2002}. Temporally dynamic representation schemes within LCS represent a potentially important approach since temporal behaviour of such kinds is viewed as a significant aspect of artificial life, biological systems, and cognition in general \citep{Ashby:1952}.   
\end{sloppypar}

In this paper we explore examples of a dynamical system representation within the XCSF learning classifier system \citep{Wilson:2001}---termed dynamical genetic programming \citep[DGP;][]{Bull:2009}. Traditional tree-based genetic programming \citep[GP;][]{Koza:1992} has been used within LCS both to calculate the action \citep{AhluwaliaBull:1999} and to represent the condition \citep[e.g.,][]{LanziPerrucci:1999}. DGP uses a graph-based representation, each node of which is constantly updated with asynchronous parallelism, and evolved using an open-ended, self-adaptive scheme. In the discrete case, each node is a Boolean function and therefore the representation is a form of random Boolean network \citep[RBN; e.g.,][]{Kauffman:1993}. In the continuous case, each node performs a fuzzy logical function and the representation is a form of fuzzy logic network \citep[FLN; e.g.,][]{KokWang:2006}. We have recently introduced the use of RBN within LCS \citep{BullPreen:2009,PreenBull:2009}. Here we extend that work to the most recent form of LCS, and to the continuous-valued domain, exploring the potential to harness the collective emergent behaviour for computation. We show that XCSF is able to solve a number of well-known immediate and delayed reward tasks using this temporally dynamic knowledge representation scheme with competitive performance with other representations. Moreover, we exploit the memory inherent to RBN for the discrete case.

The remainder of this paper is organised as follows. Section~\ref{related} provides a brief discussion of related work. Section~\ref{rbn} introduces RBN, giving a detailed description of the form and temporal behaviour. Section~\ref{xcsf} describes the XCSF learning framework. Section~\ref{ddgp} details how asynchronous RBN are here evolved and used for computation within XCSF. Section~\ref{ddgp-exp} discusses and presents results from a set of maze experiments requiring the exploitation of inherent memory within the discrete dynamical ensemble. Section~\ref{fln} introduces FLN as a continuous-valued extension of RBN and discusses the dynamical behaviour. Section~\ref{fdgp} describes how FLN are here used as continuous production system rules within XCSF. Section~\ref{fdgp-exp} presents results from experimentation with a continuous-input, discrete-action, maze environment, and a fully continuous reinforcement learning problem. Section~\ref{conclusions} provides final conclusions.

\section{Related work}
\label{related}

The most common form of discrete dynamical system is the cellular automaton \citep[CA;][]{VonNeumann:1966}, which consists of an array of cells (lattice of nodes) where the cells exist in states from a finite set and update their states with synchronous parallelism in discrete time. CA have been extensively used to model genetic regulatory networks (GRN). However, continuous network models of GRN exist which are an extension of Boolean networks, where nodes still represent genes and the connections between them regulate the influence on gene expression. Indeed, there is a growing body of work exploring the evolution of different forms of such continuous-valued GRN. Differential equations wherein gene interactions are incorporated as logical functions are a typical approach \citep[e.g.,][]{GlassKauffman:1973}. \cite{Reiter:2002} investigated the affect of the fuzzy background on the dynamics of CA with various fuzzy logic sets and found that the choice of logic used leads to significantly different behaviours. For example, applying the various logical functions to create fuzzy versions of the game of life, it was noted that certain sets of logics generated fuzzy-CA that tended toward homogeneous fuzzy behaviour, whereas others were consistent with chaotic or complex behaviour.
 
Most relevant to the form of GP used herein is the relatively small amount of prior work on graph-based representations. In particular, neural programming \citep[NP;][]{TellerVeloso:1996} uses a directed graph of connected nodes, each performing an arbitrary function. Potentially selectable functions include READ, WRITE, and IF-THEN-ELSE, along with standard arithmetic and zero-arity functions. Additionally, complex user defined functions may be used. Significantly, recursive connections are permitted and each node is executed with synchronous parallelism for some number of cycles before an output node's value is taken.

Other examples of graph-based GP typically contain sequentially updating nodes, e.g., finite state machines \citep[e.g.,][]{Fogel:1965}, cartesian GP \citep{Miller:1999}, genetic network programming \citep{Hirasawa:2001}, linear-graph GP \citep{KantschikBanzhaf:2002}, and graph structured program evolution \citep{Shirakawa:2007}. \cite{SchmidtLipson:2007} have recently demonstrated a number of benefits from graph encodings over traditional trees, such as reduced bloat and increased computational efficiency, and a significant benefit of symbolic representations is the expressive power to represent relationships between the sensory inputs \citep{Mellor:2005}.
 
\cite{Landau:2001} used a purely evolution-based form of LCS \citep[Pittsburgh style;][]{Smith:1983} in which the rules are represented as directed graphs where the genotypes are tokens of a stack-based language, whose execution builds the labeled graph. Bit-strings are used to represent the language tokens and applied to non-Markov problems. The genotype is translated into a sequence of tokens and then interpreted similarly to a program in a stack-based language with instructions to create the graph's nodes, connections and labels. Subsequently, the unused conditions and actions in the stack are added to the structure which is then popped from the stack. Tokens are used to specify the matching conditions and executable actions as well as instructions to construct the graph, and to manipulate the stack. The bit-strings were later replaced with integer tokens and again applied to non-Markov problems \citep{Landau:2005}.
 
\cite{BullOHara:2002} presented a form of fuzzy representation within LCS using radial basis function neural networks \citep[RBF;][]{MoodyDarken:1989} to embody each condition-action rule. That is, a simple class of neural-fuzzy hybrid system. Furthermore, \cite{Su:2006} explored a similar representation based on RBF within LCS. However, here the contribution of each rule is determined by its strength (which is updated by a fuzzy bucket brigade algorithm) as well as the extent to which the antecedent matches the environment. Furthermore, in contrast to \cite{BullOHara:2002}, each condition-action rule corresponds to a hidden node instead of a fully-connected network and rules are added incrementally instead of being evolved through the GA. To date, only the use of RBF has been explored as a neuro-fuzzy hybrid representation within LCS.
 
\section{Random Boolean networks}
\label{rbn}

RBN were originally introduced by Kauffman \citep[see, e.g.,][]{Kauffman:1993} to explore aspects of biological genetic regulatory networks. Since then they have been used as a tool in a wide range of areas, such as self-organisation \citep[e.g.,][]{Kauffman:1993} and computation \citep[e.g.,][]{MesotTeuscher:2005} and robotics \citep[e.g.,][]{Quick:2003}.

\begin{figure}[t]
\centering
\epsfig{file=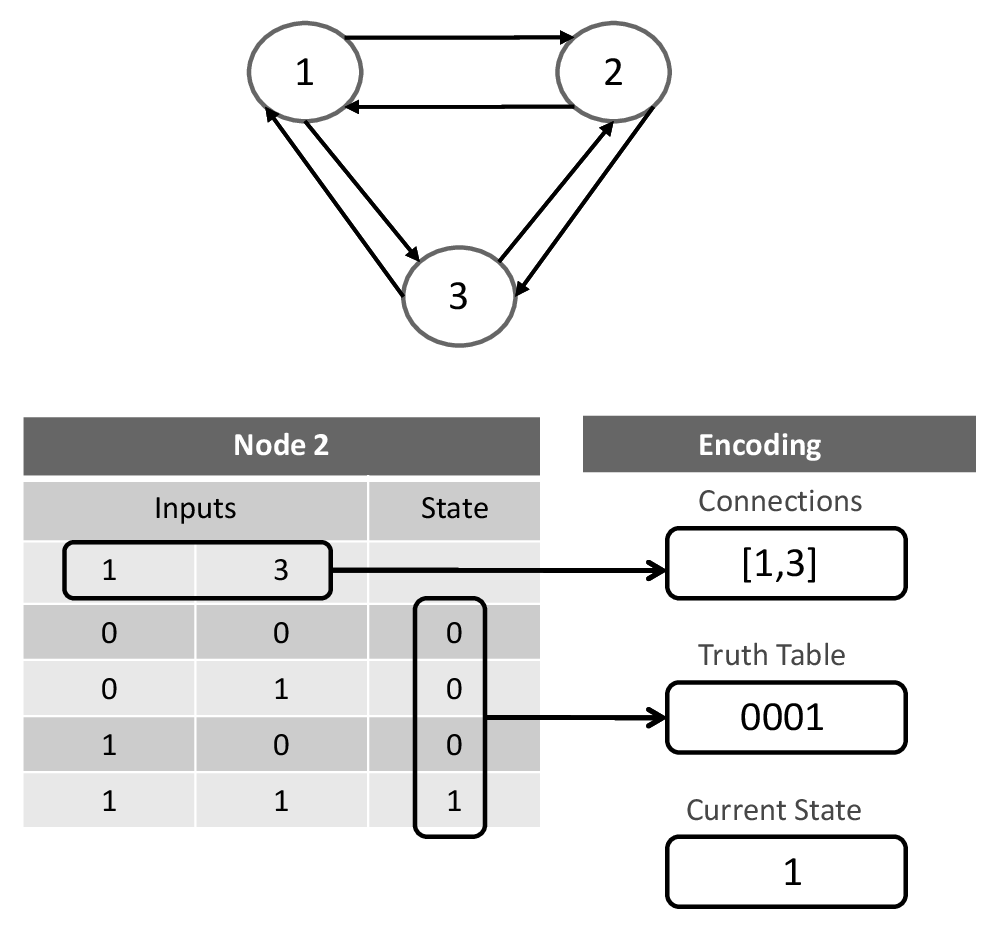,width=3.3in}
\caption{Example RBN and node encoding.}
\label{fig:ExampleRBN}
\end{figure}

An RBN typically consists of a network of $N$ nodes, each performing a Boolean function with $K$ inputs from other nodes in the network, all updating synchronously (see Figure~\ref{fig:ExampleRBN}). As such, RBN may be viewed as a generalization of binary CA and unorganized machines \citep{Turing:1948}. Since they have a finite number of possible states and they are deterministic, the dynamics of RBN eventually fall into a basin of attraction. It is well-established that the value of $K$ affects the emergent behaviour of RBN wherein attractors typically contain an increasing number of states with increasing $K$. Three phases of behaviour are suggested: ordered when $K=1$, with attractors consisting of one or a few states; chaotic when $K>3$, with a very large number of states per attractor; and, a critical regime around $K=2$, where similar states lie on trajectories that tend to neither diverge nor converge and 5-15\% of nodes change state per attractor cycle \citep[for discussions of this critical regime, e.g., with respect to perturbations see][]{Kauffman:1993}. Analytical methods have been presented by which to determine the typical time taken to reach a basin of attraction and the number of states within such basins for a given degree of connectivity $K$ \citep[see, e.g.,][]{Kauffman:1993}.

Closely akin to the work described here, \cite{Kauffman:1993} describes the use of simulated evolution to design RBN which must play a (mis)matching game wherein mutation is used to change connectivity, the Boolean functions, $K$ and $N$. He reports the typical emergence of high fitness solutions with $K$=2 to 3, together with an increase in $N$ over the initialised size. \cite{SipperRuppin:1997} extended Sipper's  heterogeneous CA approach \citep{Sipper:1997} to enable heterogeneity in the node connectivity, along with the node function; they evolved a form of RBN. \cite{VandenBroeckKawai:1990} explored the use of a simulated annealing-type approach to design feedforward RBN for the four-bit parity problem and \cite{Lemke:2001} evolved RBN of fixed $N$ and $K$ to match an arbitrary attractor.

\begin{figure}[t]
\centering
  \subfloat[Synchronous.]{\label{fig:k-affect-synch}\psfig{file=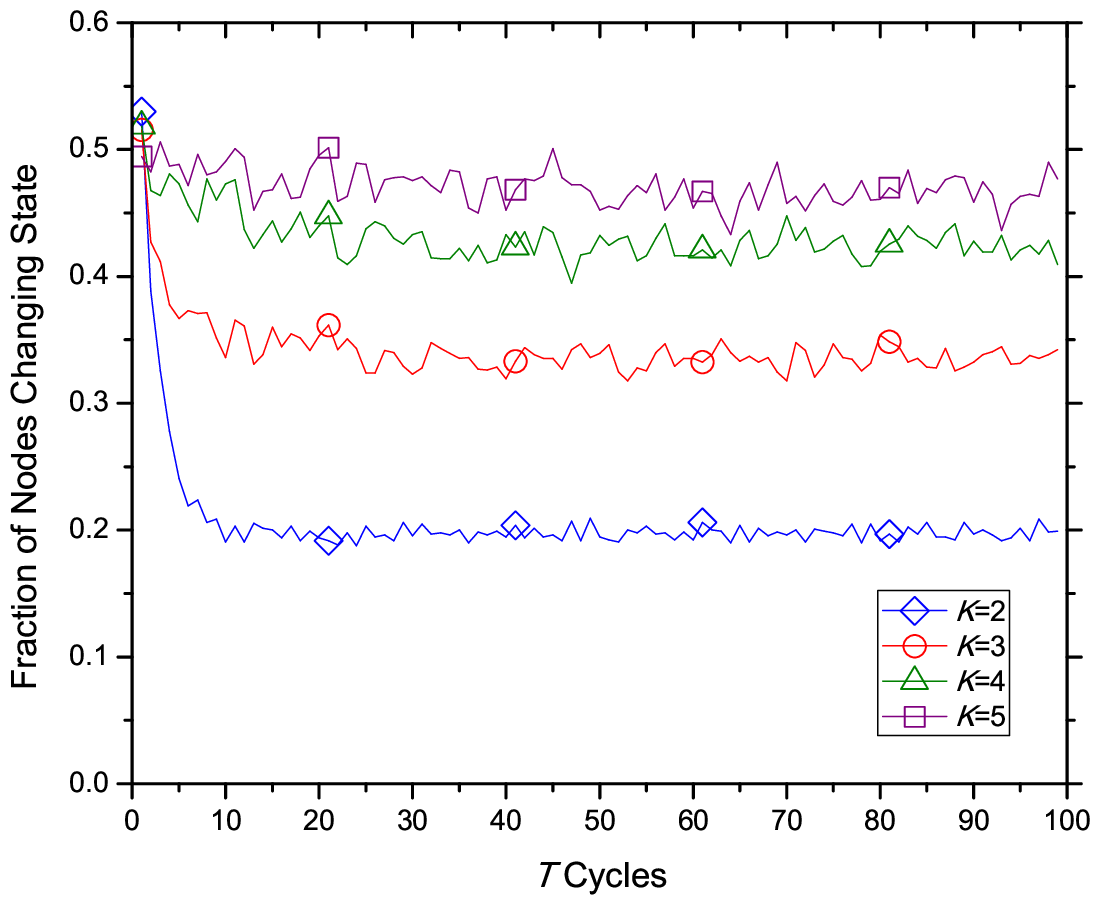,width=\figwidth}}
  \hspace{0.2in}
  \subfloat[Asynchronous.]{\label{fig:k-affect-asynch}\psfig{file=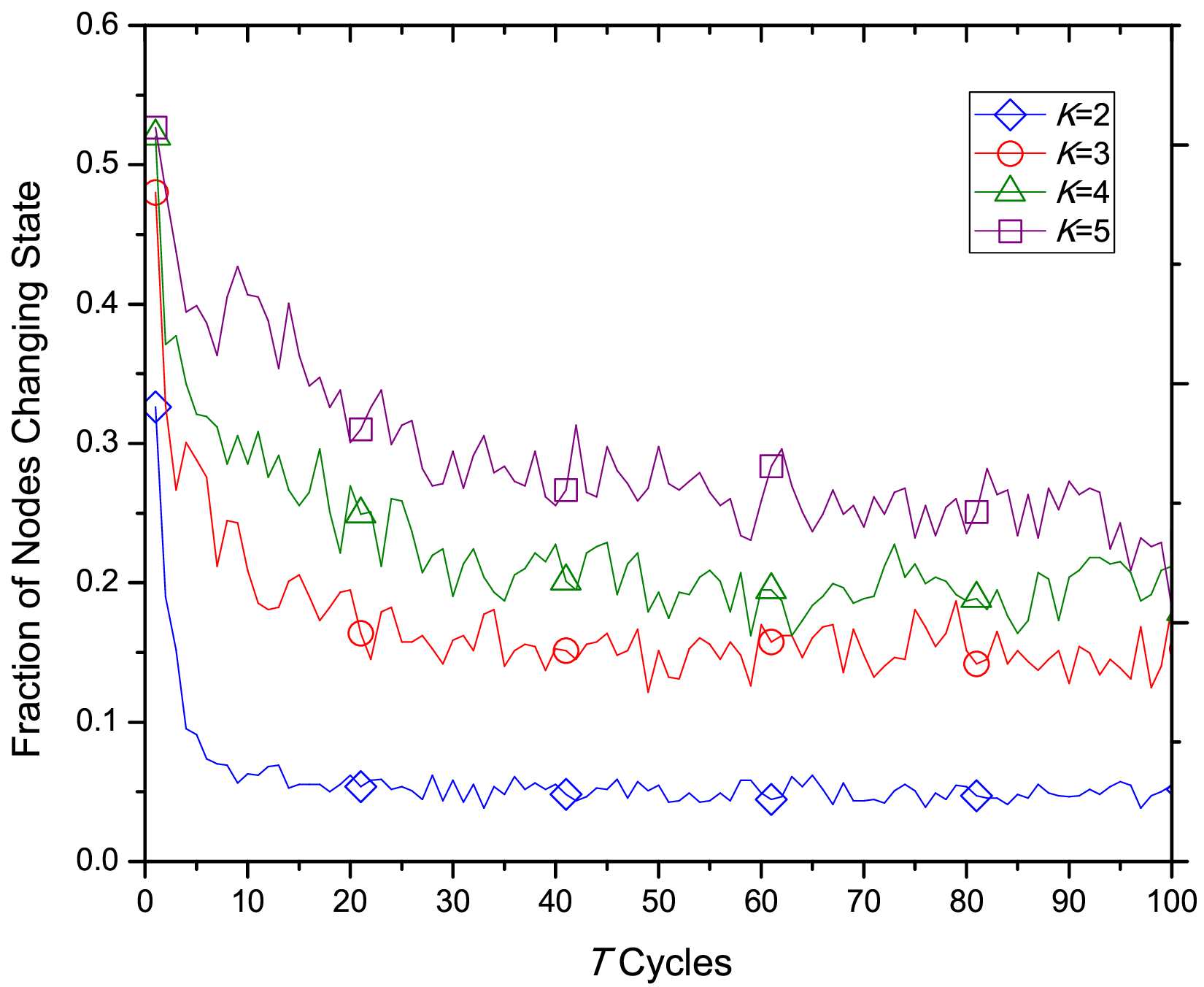,width=\figwidth}}        
\caption{The affect of $K$ on a 13 node RBN.}
\label{fig:k-affect-all}
\end{figure}

Figure~\ref{fig:k-affect-all} shows the affect of $K$ on a 13 node RBN; results are an average of 100 runs for each value of $K$. It can be seen that the higher the value of $K$, the greater the number of states the networks will cycle through, as shown by the higher rate of change of node states. Further, that after an initial rapid decline in the rate of change, this value stabilises as the states fall into their respective attractors. In the synchronous case (Figure~\ref{fig:k-affect-synch}) when $K=2$, the number of nodes changing state converges to around 20\%, and when $K=3$ to just above 35\%; thus we can see that the ordered regime occurs when approximately 20\% or less nodes are changing state each cycle, and the chaotic regime occurring for larger rates of change.

As noted above, traditional RBN consist of $N$ nodes updating synchronously in discrete time steps, but asynchronous versions have also been presented, after \cite{HarveyBossomaier:1997}, leading to a classification of the space of possible forms of RBN \citep{Gershenson:2002}. Asynchronous forms of CA have also been explored \citep[e.g.,][]{IngersonBuvel:1984} wherein it is often suggested that asynchrony is a more realistic underlying assumption for many natural and artificial systems since ``discrete time, synchronously updating networks are certainly not biologically defensible: in development the interactions between regulatory elements do not occur in a lock-step fashion'' \citep{Wuensche:2004}.

\begin{sloppypar}
Asynchronous logic devices are known to have the potential to consume less power and dissipate less heat \citep{WernerAkella:1997}, which may be exploitable during efforts towards hardware implementations of such systems. Asynchronous logic is also known to have the potential for improved fault tolerance, particularly through delay insensitive schemes \citep[e.g.,][]{DiLala:2007}. This may also prove beneficial for hardware implementations.
\end{sloppypar}

\cite{HarveyBossomaier:1997} showed that asynchronous RBN exhibit either point attractors, as seen in asynchronous CAs, or loose attractors where ``the network passes indefinitely through a subset of its possible states'' (as opposed to distinct cycles in the synchronous case). Thus the use of asynchrony represents another feature of RBN with the potential to significantly alter their underlying dynamics thereby offering another mechanism by which to aid the simulated evolutionary design process for a given task. \cite{DiPaulo:2001} showed it is possible to evolve asynchronous RBN which exhibit rhythmic behaviour at equilibrium. Asynchronous CAs have also been evolved \citep[e.g.,][]{SipperRuppin:1997}.

Figure~\ref{fig:k-affect-asynch} shows the percentage of nodes changing state on each cycle for various values of $K$ on a 13 node asynchronous RBN. It can be seen that, similar to the synchronous case (see Figure~\ref{fig:k-affect-synch}), the higher the value of $K$, the greater the number of states the networks will cycle through in an attractor. These values are significantly lower than in the synchronous case however. For example, when $K=2$, approximately 20\% of nodes change each synchronously updated cycle compared with 5\% when updated asynchronously. The difference is to be expected because, in the asynchronous case, ``the lack of synchronicity increases the complexity of the RBN, enhancing the number of possible states and interactions. And this complexity changes the attractor basins, transforming and enlarging them. This reduces the number of attractors and states in attractors'' \citep{Gershenson:2002}. As previously mentioned, in the asynchronous case there are no cycle attractors, only point and loose attractors.

\vspace{-0.2in}

\section{XCSF overview}
\label{xcsf}

An LCS rule (also termed a classifier) traditionally takes the form of an environment string consisting of the ternary alphabet [0,1,\#], a binary action string, and subsequent information including the classifier's expected payoff (reward) $P$, the error rate $\epsilon$ (in units of payoff predicted), and the fitness $f$. The \# symbol in the environment condition provides a mechanism to generalise the inputs received by matching for both logical 0 and 1 for that bit.

For each phase in the learning cycle, a match set [M] is generated from the population set [P], composed of all of the classifiers whose environment condition matches the current environmental input. In the event that the number of actions present in [M] is less than a threshold value, $\theta_{mna}$, covering is used to produce a classifier that matches the current environment state along with an action assigned randomly from those not present in [M]; typically $\theta_{mna}$ is set to the maximum number of possible actions so that there must be at least one classifier representing each action present.

Subsequently, a system prediction is made for each action in [M], based upon the fitness-weighted average of all of the predictions of the classifiers proposing the action. If there are no classifiers in [M] advocating one of the potential system actions, covering is invoked to generate classifiers that both match the current environment state and advocate the relevant action. An action is then selected using the system predictions, typically by alternating exploring (by either roulette wheel or random selection) and exploiting (the best action). In multistep problems a biased selection strategy is often employed wherein exploration is conducted at probability $p_{explr}$ otherwise exploitation occurs \citep{Lanzi:1999a}. An action set [A] is then built composed of all the classifiers in [M] advocating the selected action. Next, the action is executed in the environment and feedback is received in the form of a payoff, $P$.

In a single-step problem, [A] is updated using the current reward. The GA is then run in [A] if the average time since the last GA invocation is greater than the threshold value, $\theta_{GA}$. When the GA is run, two parent classifiers are chosen (typically by roulette wheel selection) based on fitness. Offspring are then produced from the parents, usually by use of recombination and mutation. Typically, the offspring then have their payoff, error, and fitness set to the average of their parents'. If subsumption is enabled and the offspring are subsumed by either parent, it is not included in [P]; instead the parents' numerosity is incremented. In a multistep problem, the previous action set [A]$_{-1}$ is updated using a Q-learning \citep{Watkins:1989} type algorithm and the GA may be run as described above on [A]$_{-1}$ as opposed to [A] for single-step problems. The sequence then loops until it is terminated after a predetermined number of problem instances.

In XCSF each classifier also maintains a vector of a series of weights, where there are as many weights as there are inputs from the environment, plus one extra, $x_{0}$. That is, each classifier maintains a prediction ($cl.p$) which is calculated as a product of the environmental input ($s_t$) and the classifier weight vector ($w$):
\begin{equation}
cl.p(s_t) = cl.w_0 \times x_0 + \sum\limits_{i>0} cl.w_i \times s_t(i)
\end{equation}

Each of the input weights is initially set to zero, and subsequently adapted to accurately reflect the prediction using a modified $\Delta$ rule \citep{Mitchell:1997}. The $\Delta$ rule was modified such that the correction for each step is proportional to the difference between the current and correct prediction, and controlled by a correction rate, $\eta$. The modified $\Delta$ rule for the reinforcement update is thus:
\begin{equation}
\Delta w_i = \frac{\eta}{|s_t(i)|^2} (P-cl.p(s_t))s_t(i)
\end{equation}
Where $\eta$ is the correction rate and $|s_t|^2$ is the norm of the input vector $s_t$. The values $\Delta w_i$ are used to update the weights of the classifier $cl$ with:
\begin{equation}
cl.w_i \leftarrow cl.w_i + \Delta w_i
\end{equation}

Subsequently, the prediction error $\epsilon$ is updated with:
\begin{equation}
cl.\epsilon \leftarrow cl.\epsilon + \beta(|P-cl.p(s_t)|-cl.\epsilon)
\end{equation}

This enables a more accurate, piecewise-linear, approximation of the payoff (or function), as opposed to a piecewise-constant approximation, and can also be applied to binary problems such as the Boolean multiplexer and maze environments, resulting in faster convergence to optimality as well as a more compact rule-base \citep{LoiaconoLanzi:2008}. See \cite{Wilson:2002} for further details.

\section{Discrete DGP-XCSF}
\label{ddgp}

\begin{figure*}[tbhp]
\centering
\epsfig{file=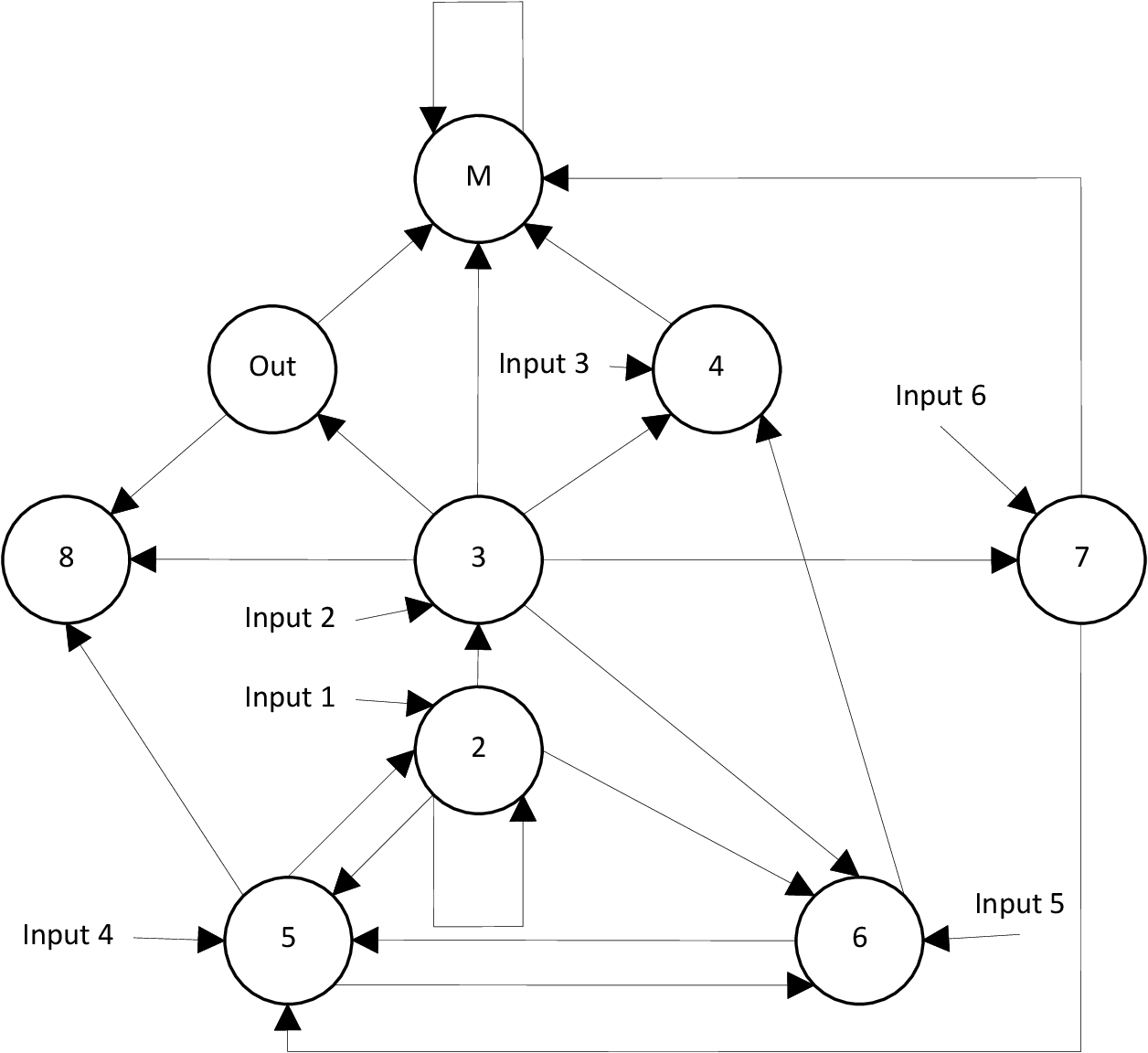,width=4.3in}
\vspace{0.2in}

\begin{tabular}{l l l}
\multicolumn{3}{l}{Prediction 1000. Error: 0.0. Accuracy: 1.0.}\\
\multicolumn{3}{l}{Experience: 822. GASetSize: 70.1. GATimeStamp: 99947}\\ \\
& Truth Table: & Connections: \\
Node 0 (M): & 10011000100000001110011010101000 & 7, 4, 0, 3, 1 \\
Node 1 (out): & 10 & 3 \\
Node 2 (I): & 00011111 & {\em Input1}, 2, 5 \\
Node 3 (I): & 0001 & {\em Input2}, 2 \\
Node 4 (I): & 11101110 & {\em Input3}, 6, 3 \\
Node 5 (I): & 0110110100001010 & {\em Input4}, 2, 7, 6 \\
Node 6 (I): & 0001011101010101 & {\em Input5}, 5, 2, 3 \\
Node 7 (I): & 0100 & {\em Input6}, 3 \\
Node 8 (N): & 00010111 & 3, 1, 5 \\
\end{tabular}

\caption{An evolved asynchronous dDGP-XCS 6-bit multiplexer rule.}
\label{fig:dDGP-examplerule}
\end{figure*}

To use asynchronous RBN as the rules within XCSF (see example rule in Figure~\ref{fig:dDGP-examplerule}), the following scheme is adopted. Each of an initial randomly created rule's nodes has $K$ randomly assigned connections, here $1 \leq K \leq 5$. There are initially as many nodes $N$ as input fields $I$ for the given task and its outputs $O$, plus one other, as will be described, i.e., $N=I+O+1$. The first connection of each input node is set to the corresponding locus of the input message. The other connections are assigned at random within the RBN as usual. In this way, the current input state is always considered along with the current state of the RBN itself per network update cycle by such nodes. Nodes are initialised randomly each time the network is run to determine [M], etc. The population is initially empty and covering is applied to generate rules as in the standard XCSF approach.

Matching consists of executing each rule for $T$ cycles based on the current input. The value of $T$ is chosen to be a value typically within the basin of attraction of the RBN. Asynchrony is here implemented as a randomly chosen node being updated on a given cycle, with as many updates per overall network update cycle as there are nodes in the network before an equivalent cycle to one in the synchronous case is said to have occurred. See \cite{Gershenson:2002} for alternative schemes.

In this study, where well-known maze problems are explored there are eight possible actions and accordingly three required output nodes. An extra matching node is also required to enable RBNs to (potentially) only match specific sets of inputs. If a given RBN has a logical `0' on the match node, regardless of its output node's state, the rule does not join [M]. This scheme has also been exploited within neural LCS \citep{Bull:2002}. A windowed approach is utilised where the output is decided by the most common state over the last $W$ steps up to $T$. For example, if the last few states on a node updating prior to cycle $T$ is 0101001 and $W=3$, then the ending node’s state would be `0' and not `1'.

\begin{sloppypar}
When covering is necessitated, a randomly constructed RBN is created and then executed for $T$ cycles to determine the status of the match and output nodes. This procedure is repeated until an RBN is created that matches the environment state.
\end{sloppypar}

Self-adaptive mutation was first applied within LCS by \cite{Bull:2000} where each rule maintains its own mutation rate $\mu$. This is similar to the approach used in evolution strategies \citep[ES;][]{Schwefel:1981} where the mutation rate is a locally evolving entity in itself, that is, it adapts during the search process. Self-adaptive mutation not only reduces the number of hand-tunable parameters of the evolutionary algorithm, it has also been shown to improve performance.

Following \cite{BullHurst:2003}, mutation only is used here. A node's truth table is represented by a binary string and its connectivity by a list of $K$ integers in the range $[1, N]$. Since each node has a given fixed $K$ value, each node maintains a binary string of length $2^K$ which forms the entries in the look-up table for each of the possible $2^K$ input states of that node, that is, as in the aforementioned work \citep{Packard:1988} on evolving CAs, for example. These strings are subjected to mutation on reproduction at the self-adapting rate $\mu$ for that rule. Hence, within the RBN representation, evolution can define different Boolean functions for each node within a given network rule, along with its connectivity map. Specifically, each rule has its own mutation rate stored as a real number and initially seeded uniform randomly in the range $[0.0,1.0]$. This parameter is passed to its offspring. The offspring then applies its mutation rate to itself using a Gaussian distribution, i.e., $\mu' = \mu e^{N(0,1)}$, before mutating the rest of the rule at the resulting rate. Due to the need for a possible different number of nodes within the rules for a given task, the DGP scheme is also of variable length. Once the truth table and connections have been mutated, a new randomly connected node is either added or the last added node is removed with the same probability $\mu$. The latter case only occurs if the network currently consists of more than the initial number of nodes. In addition, each rule maintains its own $T$ value which is initially seeded randomly between 1 and 50. Thereafter, offspring potentially increment or decrement $T$ by 1 at probability $\mu$. $W$ is evolved in a similar fashion, however it is initially seeded between 0 and $T$, and cannot be greater than $T$. Thus DGP is temporally dynamic both in the search process and the representation scheme.

Whenever an offspring classifier is created and no changes occur to its RBN when undergoing mutation, the parent's numerosity is increased and mutation rate set to that of the offspring.

\section{Discrete DGP-XCSF experimentation}
\label{ddgp-exp}

The simplest form of short-term memory is a fixed length buffer containing the $n$ most recent inputs; a common extension is to then apply a kernel function to the buffer to enable non-uniform sampling of the past values, e.g. an exponential decay of older inputs \citep{Mozer:1994}. However it is not clear that biological systems make use of such shift registers. Registers require some interface with the environment which buffers the input so that it can be presented simultaneously. They impose a rigid limit on the duration of patterns, defining the longest possible pattern and requiring that all input vectors be of the same length. Furthermore, such approaches struggle to distinguish relative temporal position from absolute temporal position \citep{Elman:1990}.

Whereas many GP systems are expression based, some have also utilised a form of memory or state. For example, linear GP \citep{Banzhaf:1997}; indexed memory \citep[e.g.,][]{Teller:1994,Brave:1996,Angeline:1997}; and work on evolving data structures which maintain internal state \citep[e.g.,][]{Langdon:1998}. In addition, some systems have used (instead of evolved) data structures to manipulate the internal state \citep[e.g., PushGP;][]{SpectorRobinson:2002}. Recently, \cite{Poli:2009} explored the use of soft assignment and soft return operations as forms of memory within linear and tree-based GP.

Here we explore and extend the hypothesis of inherent content-addressable memory existing within synchronous RBN due to different possible routes to a basin of attraction \citep{Wuensche:2004} for the asynchronous case by maintaining the node states across each input-update-output cycle. A significant advantage of this approach is that each rule/network's short-term memory is variable-length and adaptive, that is, the networks can adjust the memory parameters, selecting within the limits of the capacity of the memory, what aspects of the input sequence are available for computing predictions \citep{Mozer:1994}. In addition, as we use open-ended evolution, the maximum size of the short-term memory is also open-ended, increasing as the number of nodes within the network grows.

Here, nodes are initialised at random for the initial random placing in the maze but thereafter they are not reset for each subsequent matching cycle. Consequently, each network processes the environmental input and the final node states then become the starting point for the next processing cycle, whereupon the network receives the new environmental input and places the network on a trajectory toward a (potentially) different locally stable limit point. Therefore, a network given the same environmental input (that is, the agent's current maze perception) but with different initial node states (representing the agent's history through the maze) may fall into a different basin of attraction (advocating a different action). Thus the rules' dynamics are (potentially) constantly affected by the inputs as the system executes.

We now apply dDGP-XCSF to two well-known multistep non-Markov maze environments that require memory to resolve perceptual aliasing: Woods101 (see Figure~\ref{fig:Woods101}) and Woods102 (see Figure~\ref{fig:Woods102}).

Each cell in the maze environments is encoded with two binary bits, where white space is represented as a `*', obstacles as `O', and food as `F'. Furthermore, actions are encoded in binary as shown in Figure~\ref{fig:maze-encoding}. The task is simply to find the shortest path to the food (F) given a random start point. Obstacles (O) represent cells which cannot be occupied. In Woods1 the optimal number of steps to the food is 1.7, in maze4 optimal is 3.5 steps, in Woods101 it is 2.9, and in Woods102 it is 3.23. A teletransportation mechanism is employed whereby a trial is reset if the agent has not reached the goal state within 50 discrete movements.

\begin{figure}[t]
\fontsize{12}{15}\selectfont
\centering
\subfloat[Woods101 environment. Optimal number of steps is 2.9.]
{ 
\centering
  \begin{tabular}{ |*{7}{@{\hspace{1mm}}c@{\hspace{1mm}}| } }
    \hline
	O & O & O & O & O & O & O \\ \hline
	O & * & * & * & * & * & O \\ \hline
	O & * & O & * & O & * & O \\ \hline
	O & * & O & F & O & * & O \\ \hline
    	O & O & O & O & O & O & O \\
    \hline
  \end{tabular}
\label{fig:Woods101}
}
\\
\vspace{3mm}
\subfloat[Woods102 environment. Optimal number of steps is 3.23.]
{ 
\centering
  \begin{tabular}{ |*{7}{@{\hspace{2mm}}c@{\hspace{2mm}}| } }
    \hline
	O & O & O & O & O & O & O \\ \hline
	O & * & O & F & O & * & O \\ \hline
	O & * & O & * & O & * & O \\ \hline
	O & * & * & * & * & * & O \\ \hline
	O & * & O & * & O & * & O \\ \hline
	O & O & O & O & O & O & O \\ \hline
	O & * & O & * & O & * & O \\ \hline
	O & * & * & * & * & * & O \\ \hline
	O & * & O & * & O & * & O \\ \hline
	O & * & O & F & O & * & O \\ \hline
    	O & O & O & O & O & O & O \\
    \hline
  \end{tabular}
\label{fig:Woods102}
}
\\
\vspace{3mm}
\subfloat[Maze encoding.]
{
\centering
  \begin{tabular}{ | c | c | c@{\hspace{2mm}} | c | c | c |}
    \cline{1-2} \cline{4-6}
	Cell & Binary & & \multicolumn{3}{| c |}{Actions} \\ \cline{1-2} \cline{4-6}
	* & 00 & & 111 & 000 & 001 \\ \cline{1-2} \cline{4-6}
	O & 01 & & 110 & & 010 \\ \cline{1-2} \cline{4-6}
	F & 11 & & 101 & 100 & 011 \\ \cline{1-2} \cline{4-6}
  \end{tabular}
\label{fig:maze-encoding}
}
\caption{Experimental maze environments and encoding.}
\label{fig:mazes}
\end{figure}

\subsection{Woods101}
\begin{sloppypar}
The Woods101 maze (see Figure~\ref{fig:Woods101}) is a non-Markov environment containing two communicating aliasing states, that is, two positions which border on the same non-aliasing state and are identically sensed, but require different optimal actions. Thus, to solve this maze optimally, a form of memory must be utilised (with at least two internal states). Optimal performance has previously been achieved in Woods101 through the addition of a memory register mechanism in LCS \citep{LanziWilson:2000}, by a Corporate LCS using rule-linkage \citep{Tomlinson:2001}, and by a neural LCS using recurrent links \citep{BullHurst:2003}. Furthermore, in a proof of concept experiment, the cyclical directed graph from NP has been shown capable of representing rules with memory to solve Woods101, however it was only found to do so twice in fifty experiments \citep{BalanLuke:2004}.
\end{sloppypar}

\begin{figure}[!htbp]
\centering
\subfloat[Number of steps to goal (circle).] {
	\label{fig:dDGP-XCSF-Woods101-Perf} 
	\epsfig{file=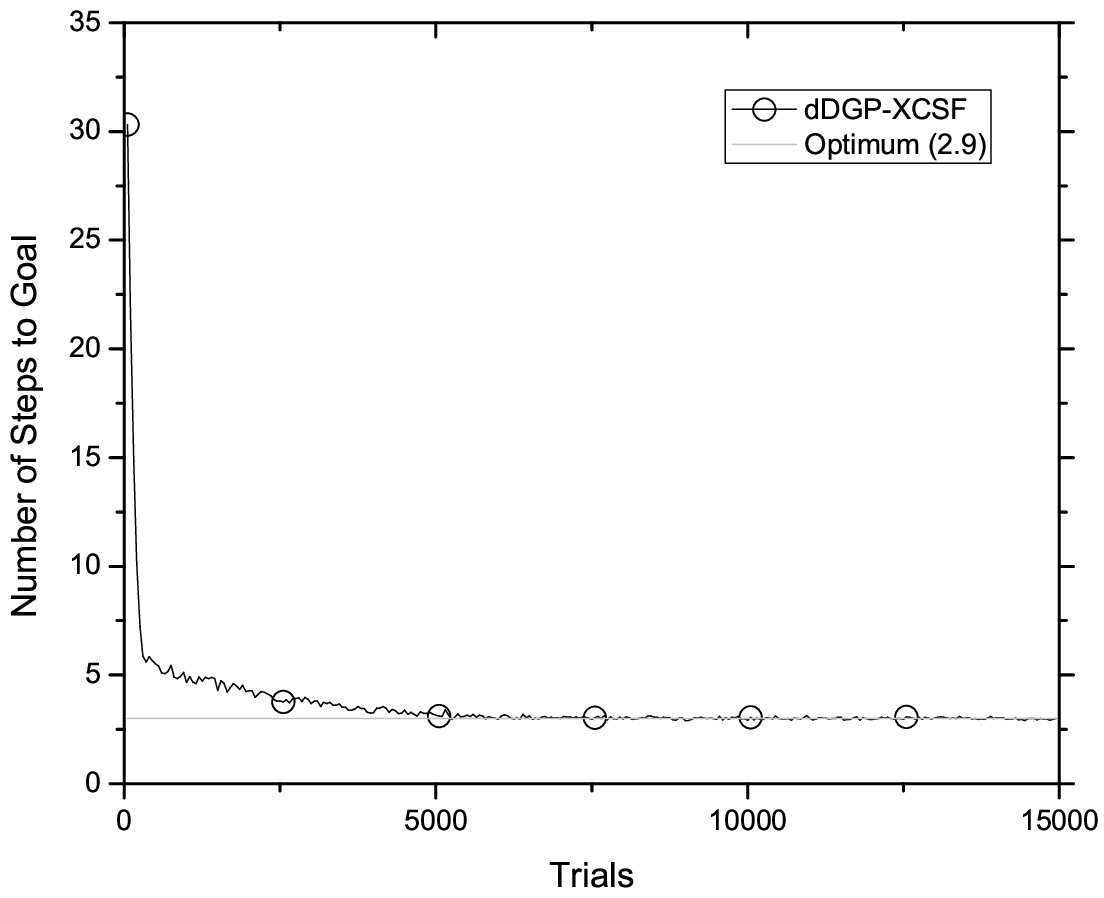,width=\figwidth} 
} \\
\subfloat[Average mutation rate (square) and number of macro-classifiers (circle).] {
	\label{fig:dDGP-XCSF-Woods101-SizeMut} 
	\epsfig{file=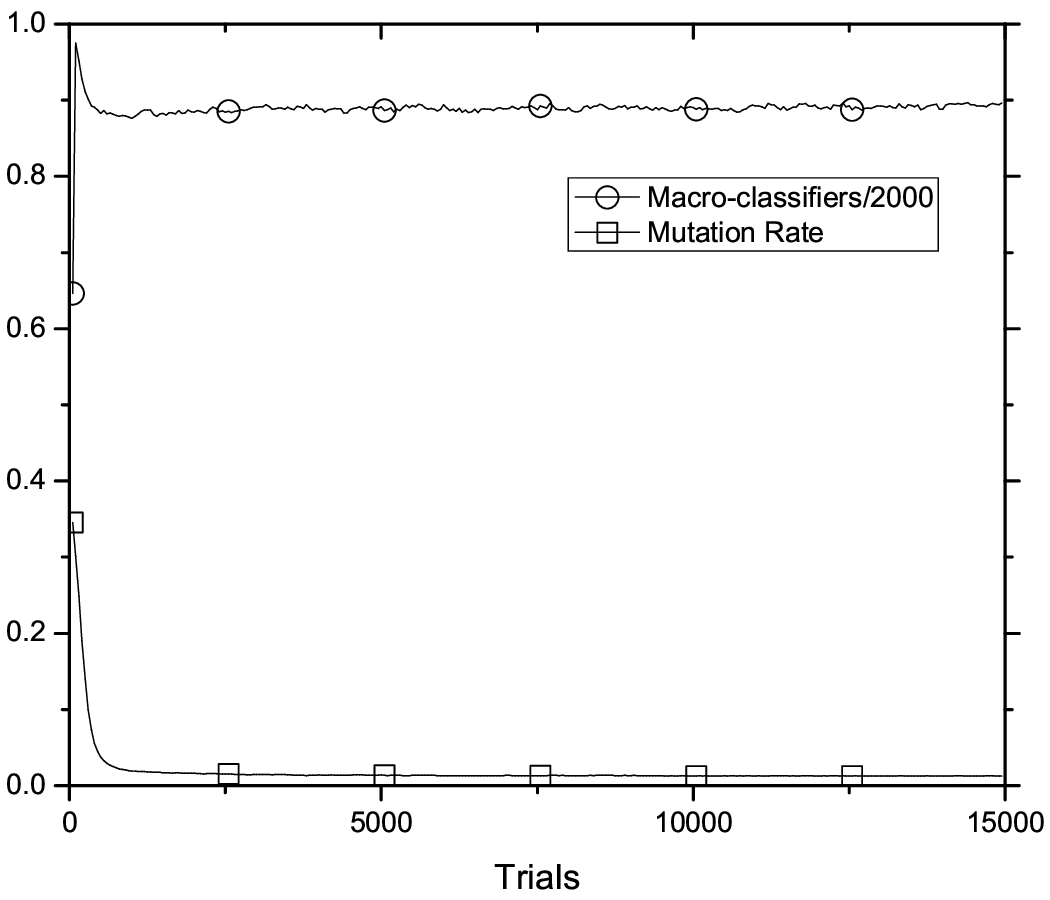,width=\figwidth} 
} \\
\subfloat[Average number of nodes (square), connections (diamond), $T$ (circle) and $W$ (triangle).] { 
	\label{fig:dDGP-XCSF-Woods101-Topology} 
	\epsfig{file=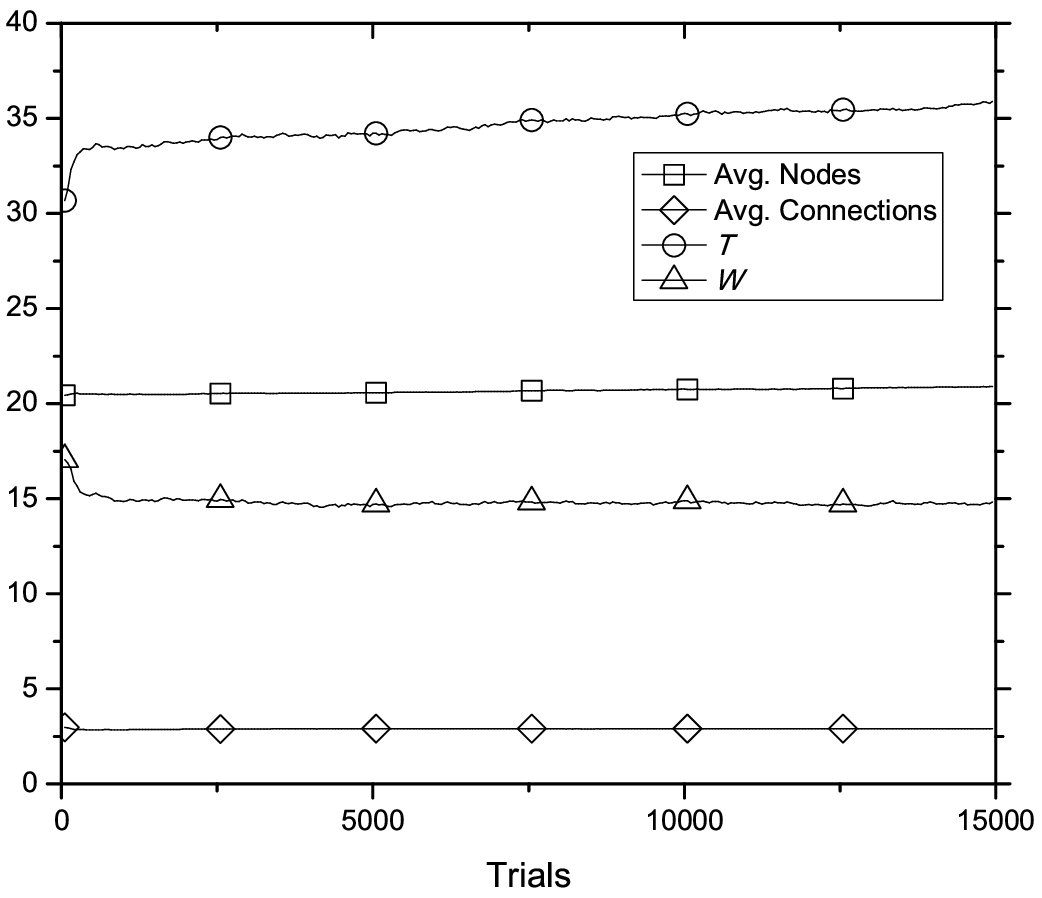,width=\figwidth} 
}
\caption{dDGP-XCSF Woods101 performance.}
\label{fig:dDGP-XCSF-Woods101-all}
\end{figure}

Figure~\ref{fig:dDGP-XCSF-Woods101-all} shows the performance of dDGP-XCSF in the Woods101 environment with $P=2000$, $\nu=5$, $\theta_{GA}=25$, $\theta_{del}=20$, $\beta=0.2$, $\eta=0.2$, $x_0=1$, $p_{expl}=1.0$, and $N_{init} = 20$ (16 inputs, 3 outputs, 1 match node). Here, optimality is observed after approximately 6,000 trials (Figure~\ref{fig:dDGP-XCSF-Woods101-Perf}). This is similar to the performance of LCS using a 1-bit memory register \citep[$\sim$7,000 trials, $P=800$;][]{LanziWilson:2000}. The number of macro-classifiers in the population converges to around 1800 (Figure~\ref{fig:dDGP-XCSF-Woods101-SizeMut}). Furthermore, the average number of nodes in the networks increases by almost one and the number of connections declines fractionally (Figure~\ref{fig:dDGP-XCSF-Woods101-Topology}). The mutation rate (also Figure~\ref{fig:dDGP-XCSF-Woods101-SizeMut}) declines rapidly from approx 35\% to its lowest point, 1.2\%, around the 6,000 trial, which is at the same moment optimal performance is also observed. Lastly, Figure~\ref{fig:dDGP-XCSF-Woods101-Topology} conveys that the first thousand trials sees a rapid increase in the number of cycles, $T$, (30.6 to 34.4) and a rapid decrease in the value of $W$ (17 to 14.7). Subsequently, $T$ continues to increase, (although at a much slower rate) along with the average number of nodes in the networks; $W$ remains stable at just fewer than 15.

\subsection{Woods102}
\begin{figure}[!htbp]
\centering
\subfloat[Number of steps to goal (circle).]
{ \label{fig:dDGP-XCSF-Woods102-Perf} \epsfig{file=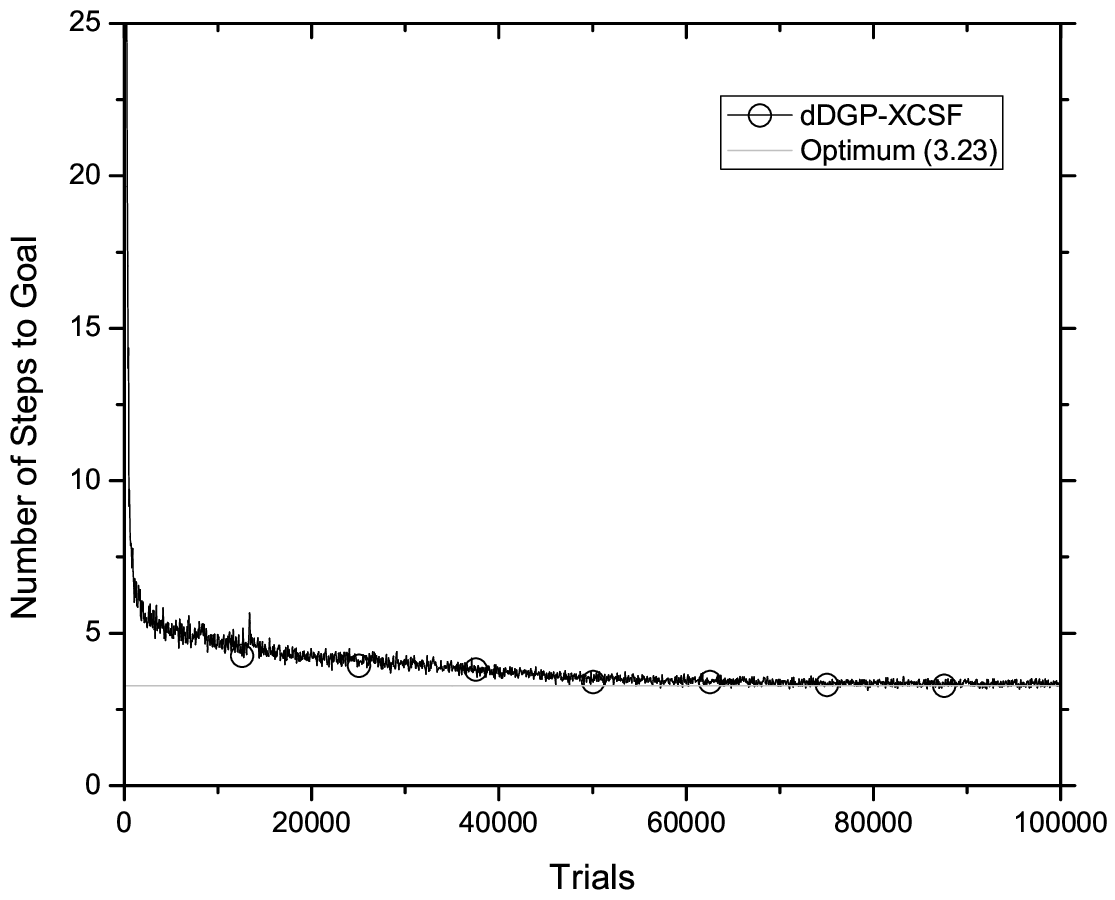,width=\figwidth} }
\\
\subfloat[Average mutation rate (square) and number of macro-classifiers (circle).]
{ \label{fig:dDGP-XCSF-Woods102-SizeMut} \epsfig{file=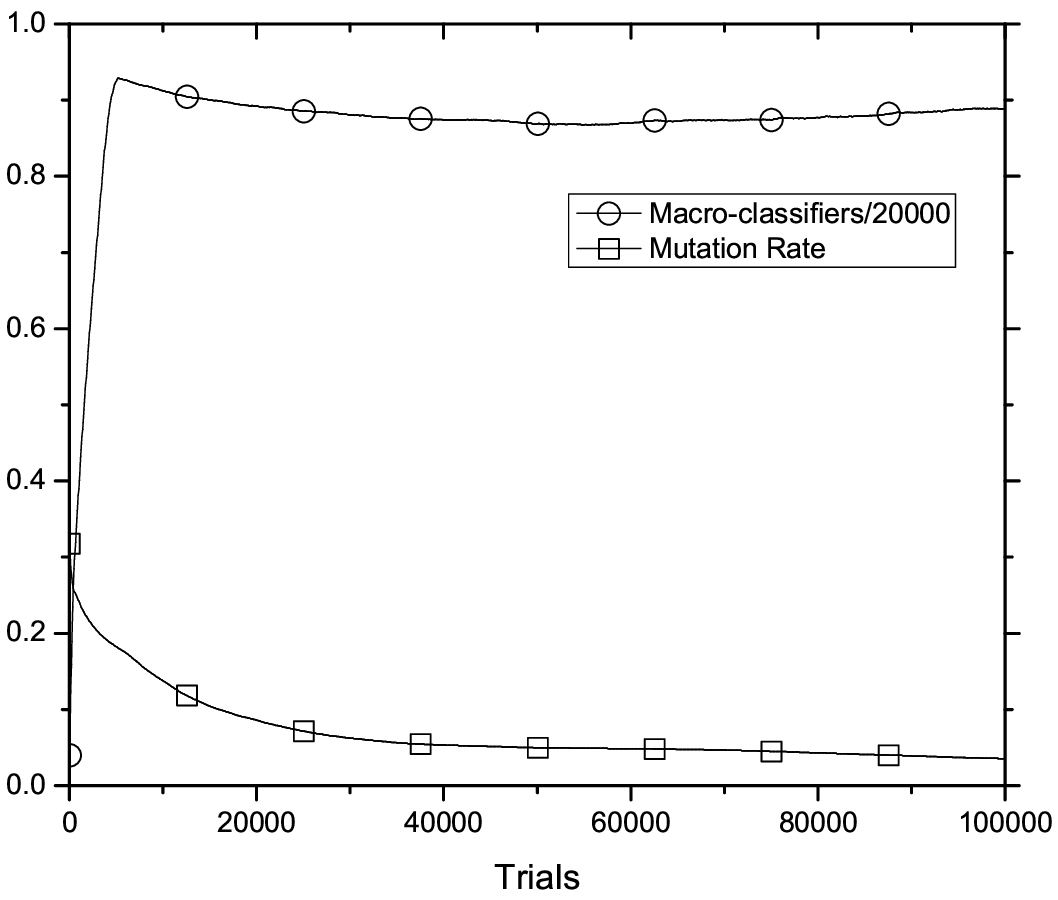,width=\figwidth} }
\hspace{0.2in}
\subfloat[Average number of nodes (square), connections (diamond), $T$ (circle) and $W$ (triangle).]
{ \label{fig:dDGP-XCSF-Woods102-Topology} \epsfig{file=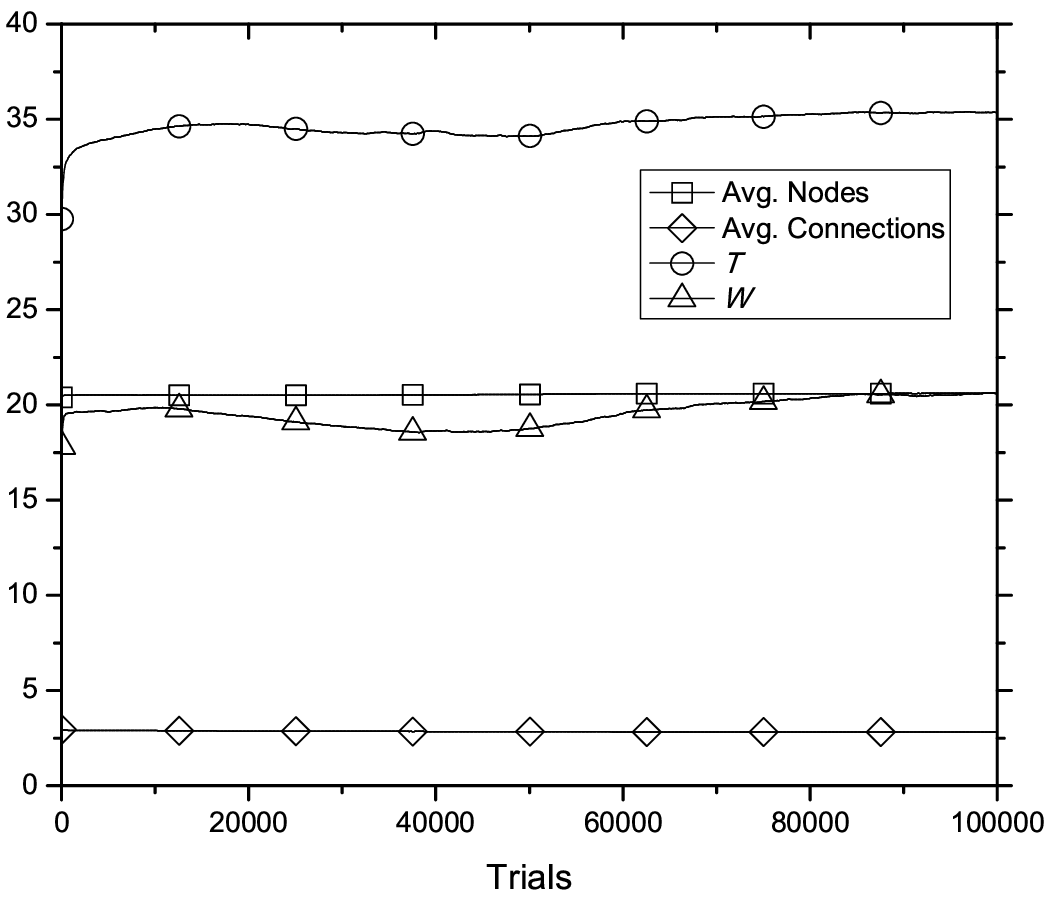,width=\figwidth} }
\caption{dDGP-XCSF Woods102 performance.}
\label{fig:dDGP-XCSF-Woods102-all}
\end{figure}

The Woods102 maze (see Figure~\ref{fig:Woods102}) is a non-Markov environment containing aliasing conglomerates, that is, adjacent aliasing states. The introduction of aliasing conglomerates increases the complexity of the learning task facing the agent significantly. ``It would appear that three memory-register bits are required to resolve [the] perceptual aliasing. However, since the two situations occur in separate parts of the environment, there is the possibility that an optimal policy could evolve in which certain register bits are used in more than one situation, thus requiring fewer bits in all. It is therefore not clear how large a bit-register is strictly necessary'' \citep{LanziWilson:2000}. However, in practice, register redundancy was found to be important and an 8-bit memory register was required within LCS to solve the maze optimally, with 2 and 4-bit registers achieving only 4 and 3.7 steps respectively \citep{LanziWilson:2000}. Figure~\ref{fig:dDGP-XCSF-Woods102-all} shows the performance of dDGP-XCSF in Woods102 with the same parameters used in the prior experiment, however, here $p_{expl}=0.1$ and $P=20,000$. Although a population size of 20,000 may seem disproportionate, a population of 2,000 classifiers was required for Woods101, representing a scale up of $10\times$, which can be compared with the increase required by LCS with a memory register (800 to 6,000, or $7.5\times$), where the potential number of internal actions required rises from $3^{1}=3$ to $3^{8}=6561$ \citep{LanziWilson:2000}, thus resources are clearly not increasing as quickly as the search space.

Optimality is observed after approximately 80,000 trials (Figure~\ref{fig:dDGP-XCSF-Woods102-Perf}), this is slower than LCS with an explicit 8-bit memory register \citep[$\sim$30,000 trials, $P=6000$;][]{LanziWilson:2000}. However here the size of the memory did not need to be predetermined as it is inherent within the networks, and the action selection policy remains constant, with constant GA activity, unlike in \cite{LanziWilson:2000}. The number of macro-classifiers in the population converges to around 17,750 (Figure~\ref{fig:dDGP-XCSF-Woods102-SizeMut}). Furthermore, the average number of nodes in the networks increases fractionally to 20.6 and the number of connections declines on average from 2.95 to 2.82 (Figure~\ref{fig:dDGP-XCSF-Woods102-Topology}). The mutation rate (Figure~\ref{fig:dDGP-XCSF-Woods102-SizeMut}) declines rapidly over the first 40,000 trials from 32\% to 5\% and reaches its lowest point, 3.5\%, at 100,000 trials. Lastly, from Figure~\ref{fig:dDGP-XCSF-Woods102-Topology} it can be seen that on average $T$ increases from 30 to 35 and $W$ from 17.5 to 20.5.

\section{Fuzzy logic networks}
\label{fln}

FLN \citep{KokWang:2006,Cao:2007} can be seen as both a generalization of fuzzy-CA and RBN, where the Boolean functions from RBN are replaced with fuzzy logical functions from fuzzy set theory. Thus, FLN generalize RBN through a continuous representation and generalize fuzzy-CA through a less restricted graph topology. \cite{KokWang:2006} explored 3-gene regulation networks using FLN and found that not only were FLN able to represent the varying degrees of gene expression but also that the dynamics of the networks were able to mimic a cell's irreversible changes into an invariant state or progress through a periodic cycle.

\begin{figure}[!htbp]
\centering
\epsfig{file=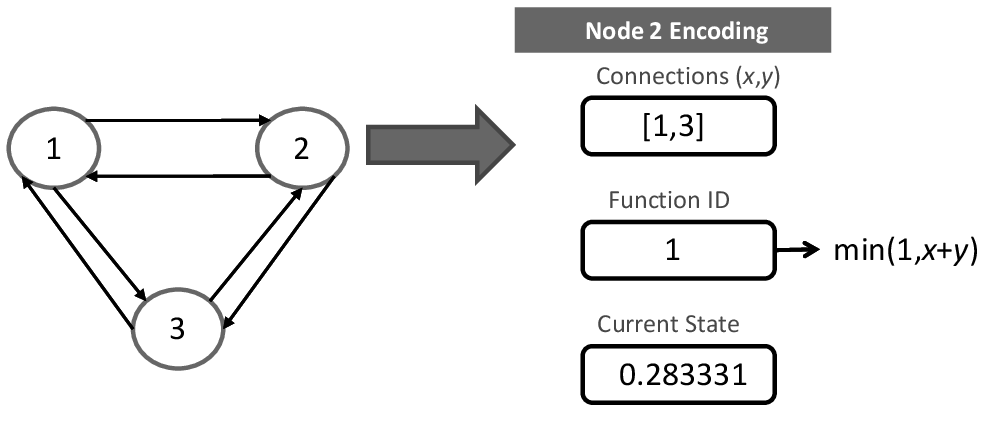,width=3.4in}
\caption{Example FLN and node encoding. Node 2 receives inputs from node 1 and 3 and performs a fuzzy OR.}
\label{fig:example-fln}
\end{figure}

FLN are defined as, given a set of $N$ variables (genes),
\begin{equation}
\begin{split}
F(t) = (F_1(t), F_2(t), ..., F_N(t)), \\
F_i(t) \in [0,1] (i=1, 2, ..., N)
\end{split}
\end{equation}
index $t$ represents time; and the variables are updated by means of dynamic equations, 
\begin{equation}
F_i(t+1) = \Lambda_i(F_{i1}(t), F_{i2}(t), ..., F_{iK}(t) 
\end{equation}
where $\Lambda_i$ is a randomly chosen fuzzy logical function. The total number of choices for fuzzy logical functions is decided only by the number of inputs. If a node has $K (1\le K \le N)$ inputs, then there are $2^K$ different fuzzy logical functions. In the definition of FLN, each node, $F_i(t)$ has $K$ inputs (see Figure~\ref{fig:example-fln}). The membership function is defined as a function $\Lambda_u : U \rightarrow [0,1]$ where $\Lambda_u$ is the degree of membership \citep{Cao:2007}. In all work thus far, all nodes are updated simultaneously, that is, synchronously.

\begin{sloppypar}
A number of different fuzzy logic sets have been introduced since the original Max/Min method was proposed. Other commonly used fuzzy logics include CFMQVS, Probabilistic, MV, and gcd/lcm \citep{Reiter:2002}. As previously mentioned, the choice of fuzzy set can result in significantly different behaviour. Therefore, in this paper a range of the most commonly used logics is potentially selectable (see Table~\ref{table:FuzzyLogics}), leaving evolution to identify the most appropriate combinations for a given problem.
\end{sloppypar}

\begin{table}[!htb]
\caption{Selectable fuzzy logic functions.}
\centering
\begin{tabular}{l l l }
\hline\hline
ID & Function & Logic \\
\hline
0 & Fuzzy OR (Max/Min) & $max(x,y)$ \\
1 & Fuzzy AND (CFMQVS) & $x \times y$ \\
2 & Fuzzy AND (Max/Min) & $min(x,y)$ \\
3 & Fuzzy OR (CFMQVS and MV) & $min(1,x+y)$  \\
4 & Fuzzy NOT & $1-x$ \\
5 & Identity & $x$ \\ [0ex]
\hline
\end{tabular}
\label{table:FuzzyLogics}
\end{table}

\begin{figure}[!htb]
\centering
\subfloat[Asynchronous.]
{ \label{fig:fuzzy-k-affect-asynch} \epsfig{file=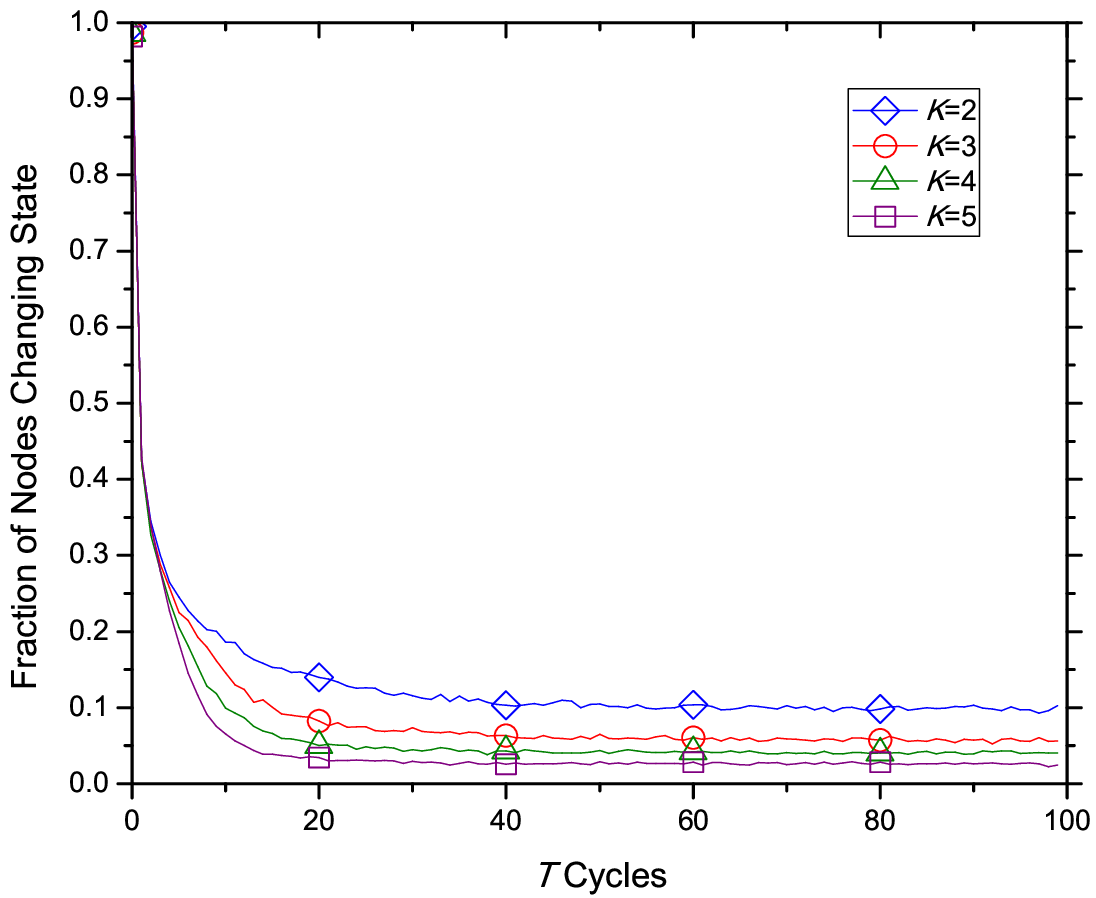,width=\figwidth} }
\hspace{0.2in}
\subfloat[Synchronous.]
{ \label{fig:fuzzy-k-affect-synch} \epsfig{file=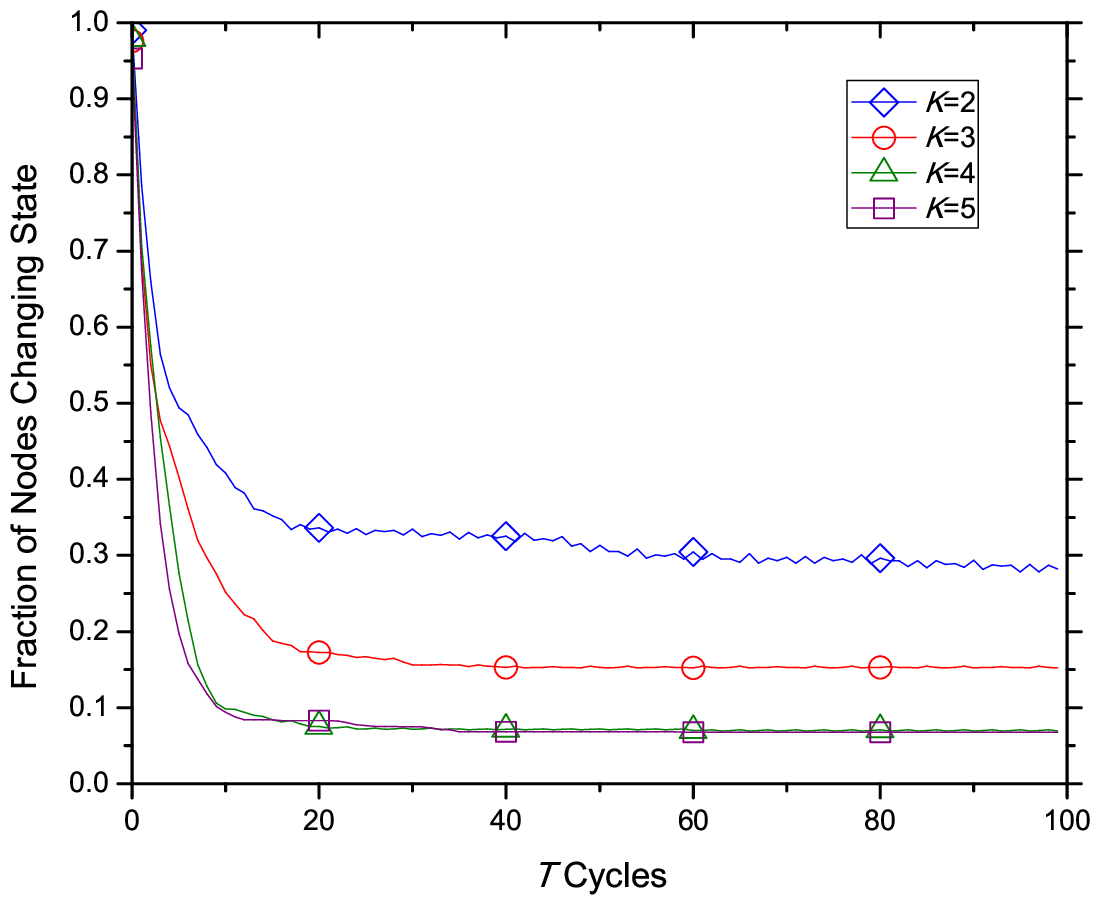,width=\figwidth} }
\caption{The affect of $K$ on a 13 node FLN.}
\label{fig:fuzzy-affect}
\end{figure}

As previously mentioned, FLN are typically updated synchronously, however asynchronous schemes in CA, RBN, and fuzzy-CA have been shown to provide a number of benefits, such as modeling the dynamics of GRN more realistically. Figure~\ref{fig:fuzzy-k-affect-asynch} shows the affect of $K$ on a 13 node FLN updated asynchronously and Figure~\ref{fig:fuzzy-k-affect-synch} when updated synchronously; results are an average of 100 experiments for each value of $K$. In contrast to RBN where larger $K$ results in an increased percentage of nodes changing state per update cycle \citep{Kauffman:1993}, it can be seen that with FLN the greater the value of $K$, the less the number of states the networks will cycle through within an attractor. This is due to the tendency of the fuzzy logic functions to gravitate to extremes (i.e., 0 or 1) with increased inputs. After an initial rapid decline in the rate of change, the networks begin to stabilise as the states fall into their respective attractors. However, similar to RBN, it can be seen that an asynchronous updating scheme results in a lower percent of nodes changing state when compared to the synchronous case. In the asynchronous case, when $K=2$, the number of nodes changing state converges to around 10\% compared with 30\% of synchronous nodes, and when $K=5$ to approximately 2.5\% compared with 7\% of nodes in the synchronous case.

\section{Fuzzy DGP-XCSF}
\label{fdgp}

Whereas FLN have been used previously to model aspects of GRN, no prior studies have explored the evolution of the networks for computation. Furthermore, all prior studies have only considered a synchronous updating scheme. To use asynchronous FLN as the rules within XCSF (hereinafter, fDGP-XCSF), the following scheme is adopted. Each of an initial randomly created rule's nodes has $K$ randomly assigned connections, here $0 \le K \le 5$, where a node with $K=0$ thus retains a constant node state. There are initially as many nodes $N$ as input fields $I$ for the given task and its outputs $O$, plus one other, for matching, i.e., $N=I+O+1$. The first connection of each input node is set to the corresponding locus of the input message. The other connections are assigned at random within the FLN. Node states are initialised at random for the first step of a trial but thereafter they are not reset for each subsequent matching cycle. The population is initially empty and covering is applied to generate rules as in the standard XCSF approach.

If a given FLN has a (real) value of fewer than 0.5 on the match node, regardless of the state of its outputs, the rule does not join [M] (see Figure~\ref{fig:example-fln}). This scheme has also been exploited within neural LCS \citep{Bull:2002}. The output nodes are discretised in a similar fashion where a state fewer than 0.5 translates to a binary 0, otherwise 1. Furthermore, a windowed approach is utilised whereby the final state of each node is calculated as an average over the last $W$ cycles to $T$.

A node's function is represented by an integer which references the appropriate operation to execute upon its received inputs (see Table~\ref{table:FuzzyLogics} for the fuzzy functions used). Further, each node's connectivity is represented as a list of $MAX\_K$ integers (here $MAX\_K = 5$) in the range $[0, N]$, where 0 represents no input to be received on that connection. Each integer in the list is subjected to mutation on reproduction at the self-adapting rate $\mu$ for that rule. Hence, within the representation, evolution can select different fuzzy logic functions for each node within a given network rule, along with its connectivity map. 

\section{Fuzzy DGP-XCSF experimentation}
\label{fdgp-exp}

\subsection{2-D continuous gridworld environment}

The 2-D continuous gridworld environment \citep{BoyanMoore:1995} is a two dimensional environment wherein the current state is a real valued coordinate $(x,y) \in [0,1]^2$. The agent is initially randomly placed within the grid and attempts to find the shortest path to the goal, located in the upper right corner; more specifically, in this paper the goal is found when $x+y>1.9$, at which point the agent is given a fixed reward of 1000, otherwise 0 is given. Any action that would take the system outside of the environment moves the system to the nearest boundary. A teletransportation mechanism is employed whereby a trial is reset if the agent has not reached the goal state within 500 movements. As actions, the agent may choose one of four possible movements (north, south, east, or west) each of which is a step size, $s$, of 0.05. The optimal number of steps is thus 18.6. The continuous state space, combined with the long sequence of actions required to reach the goal, make the Continuous Gridworld one of the most challenging multistep problems hitherto considered by LCS \citep{Lanzi:2005b}.

\begin{figure}[!htbp]
\centering
\subfloat[Performance (circle), error (square), macro-classifiers (triangle) and mutation rate (diamond).]
{ \label{fig:fDGP-grid-perf} \epsfig{file=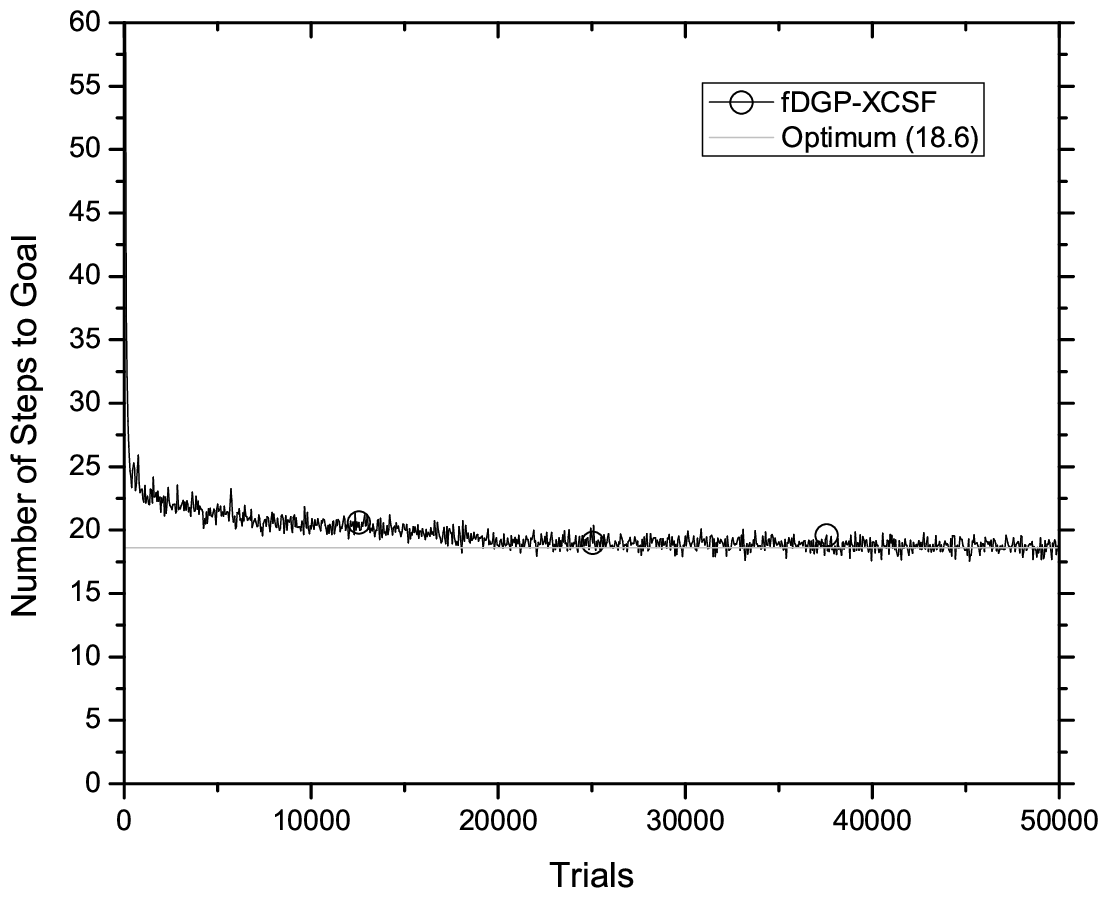,width=\figwidth} }
\\
\subfloat[Average number of macro-classifiers (circle) and mutation rate (square).]
{ \label{fig:fDGP-grid-SizeMut} \epsfig{file=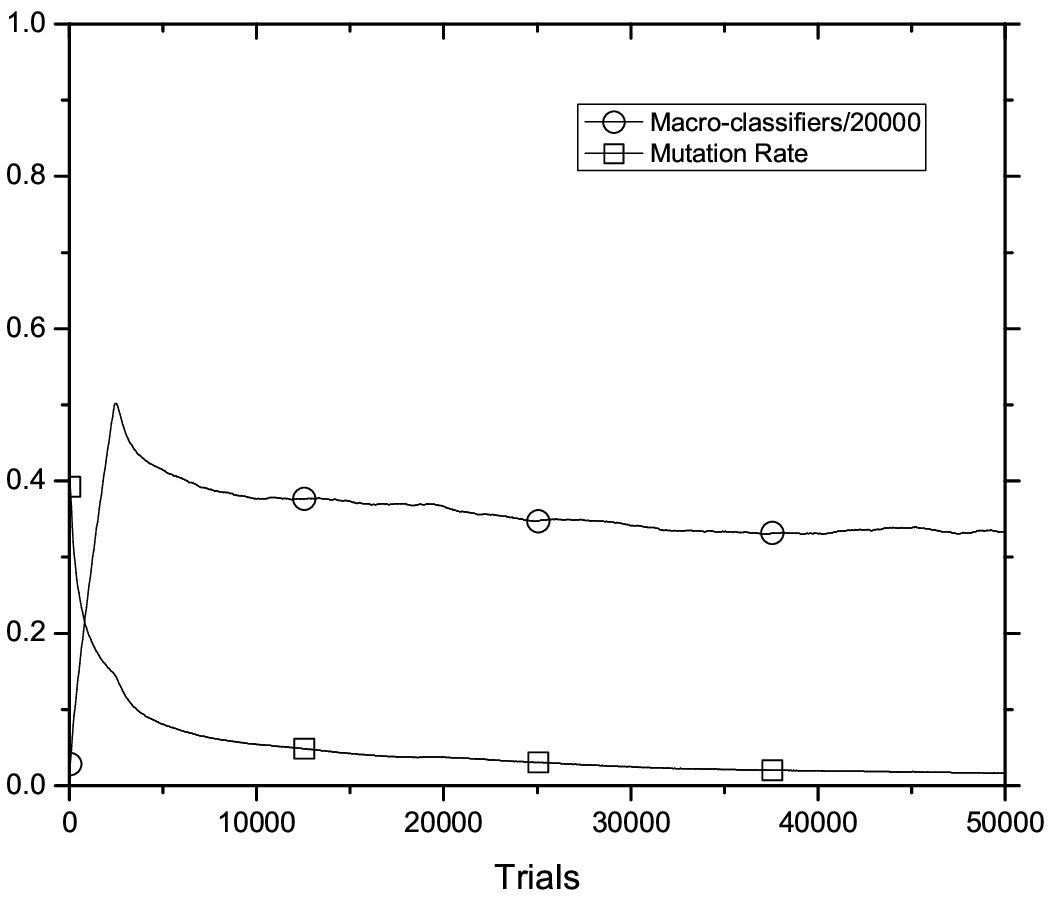,width=\figwidth} }
\hspace{0.2in}
\subfloat[Average number of nodes (square), connections (diamond), $T$ (circle) and $W$ (triangle).]
{ \label{fig:fDGP-grid-top} \epsfig{file=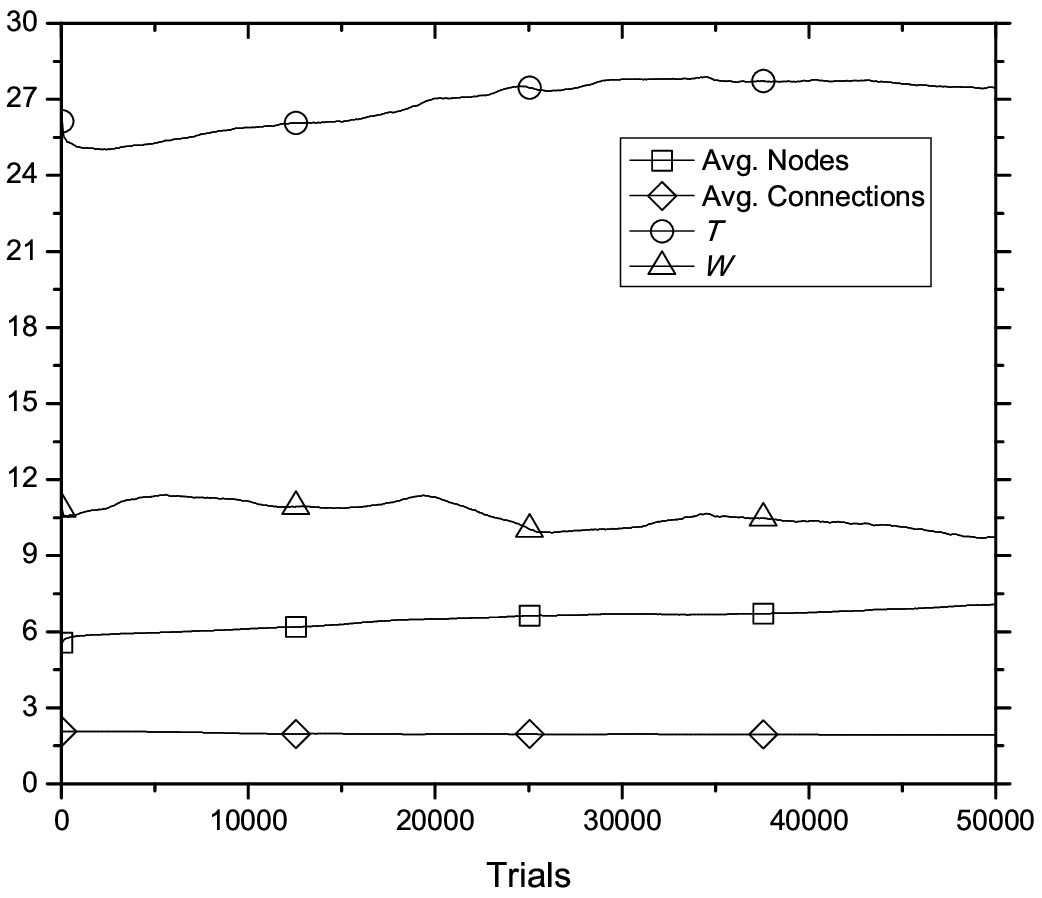,width=\figwidth} }
\caption{fDGP-XCSF continuous grid(0.05) performance.}
\label{fig:fDGP-grid-all}
\end{figure}

Figure~\ref{fig:fDGP-grid-all} shows the performance of fDGP-XCSF in the continuous gridworld environment using the same parameters used by \cite{Lanzi:2005b}. However, here $P=20,000$, $N_{init}=5$ (2 inputs, 2 outputs, 1 match node). From Figure~\ref{fig:fDGP-grid-perf} it can be seen that an optimal solution is learnt around 30,000 trials, which is slower than XCSF with interval-conditions \citep[$\sim$15,000 trials, $P=10,000$;][]{Lanzi:2005b}, however is similar in performance to an MLP-based neural-XCSF \citep{Howard:2010}. The average mutation rate within the networks (see Figure~\ref{fig:fDGP-grid-SizeMut}) declines rapidly from 40\% to 5\% after 10,000 trials and then declines at a slower rate until reaching a bottom around 2.5\% after 50,000 trials. The number of (non-unique) macro-classifiers (also Figure~\ref{fig:fDGP-grid-SizeMut}) initially grows rapidly, reaching a peak at 10,000 before declining to around 6,900. Furthermore, from Figure~\ref{fig:fDGP-grid-top} it can be seen that the average number of nodes in the FLNs increases from 5 to 7.1 and the average number of connections within the networks remains near static around 2. Additionally, the average value of $W$ remains static around 10, while the value of $T$ increases slightly, on average, from 26 to \~27.

\subsection{Continuous-action frog problem}
The frog problem \citep{Wilson:2004,Wilson:2007} is a single-step problem with a non-linear continuous-valued payoff function in a continuous one-dimensional space. A frog is given the learning task of jumping to catch a fly that is at a distance, $d$, from the frog, where $0 \le d \le 1$. The frog receives a sensory input, $x(d)=1-d$, before jumping a chosen distance, $a$, and receiving a reward based on its new distance from the fly, as given by:
\begin{equation}
P(x,a) = \left\{ \begin{array}{rl}
 x+a &\mbox{ : $x+a\le1$} \\
 2-(x+a) &\mbox{ : $x+a\ge1$}
       \end{array} \right.
\end{equation}

In the continuous-action case, the frog may select any continuous number in the range [0,1] and thus the optimal achievable performance is 100\%. 

\cite{Wilson:2007} presented a form of XCSF where the action was computed directly as a linear combination of the input state and a vector of action weights, and conducted experimentation on the continuous-action Frog problem, selecting the classifier with the highest prediction for exploitation. \cite{Tran:2007} subsequently extended this by adapting the action weights to the problem through the use of an ES. Moreover, the exploration action selection policy was modified from purely random to selecting the action with the highest prediction. After reinforcement updates and running the ES, the GA is invoked using a combination of mixed crossover and mutation. They reported greater than 99\% performance after an averaged number of 30,000 trials ($P=2000$), which was superior to the performance reported by \cite{Wilson:2007}. More recently, \cite{Ramirez-Ruiz:2008} applied a fuzzy-LCS with continuous vector actions, where the GA only evolved the action parts of the fuzzy systems, to the continuous-action frog problem, and achieved a lower error than Q-learning (discretized over 100 elements in $x$ and $a$) after 500,000 trials ($P=200$).

To accommodate continuous-actions, the following modifications were made to fDGP-XCSF. Firstly, the output nodes are no longer discretized, instead providing a real numbered output in the range [0,1]. After building [M] in the standard way, [A] is built by selecting a single classifier from [M] and adding matching classifiers whose actions are within a predetermined range of that rule's proposed action (here the range, or window size, is set to $\pm0.005$). Parameters are then updated and the GA executed as usual in [A]. Exploitation functions by selecting the single best rule from [M]; the following experiments compare the performance achieved using various criteria to select the best rule from the match set. The parameters used here are the same as used by \cite{Wilson:2004,Wilson:2007} and \cite{Tran:2007}, i.e., $P=2000$, $\nu=5$, $\theta_{GA}=48$, $\theta_{del}=50$, $\epsilon_0=0.01$, $\beta=0.2$, $\eta=0.2$, $x_0=1$. Only one output node is required and thus $N_{init}=3$.

\begin{figure}[!htb]
\centering
\subfloat[Performance (circle), error (square), macro-classifiers (triangle) and mutation rate (diamond).]
{ \label{fig:fDGP-frog-asynch} \epsfig{file=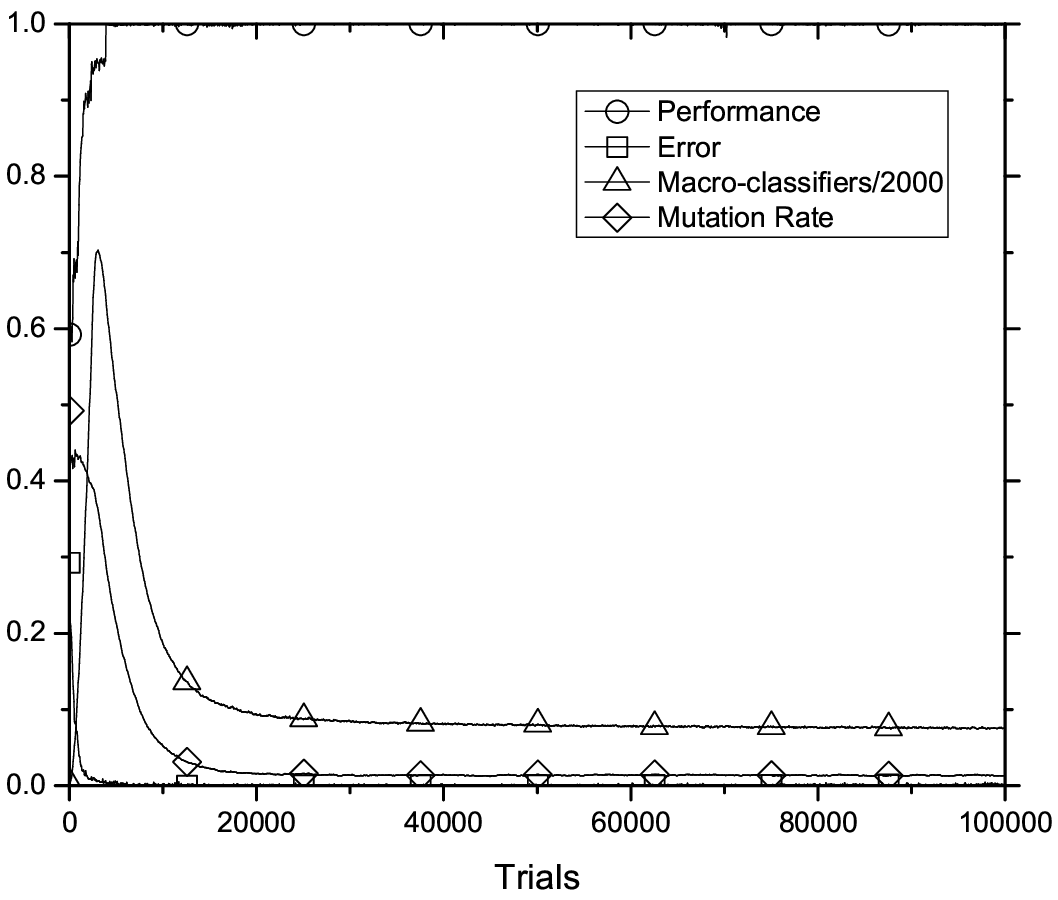,width=\figwidth} }
\hspace{0.2in}
\subfloat[Average number of nodes (triangle), average connections (square) and average $T$ (circle).]
{ \label{fig:fDGP-frog-asynch-top} \epsfig{file=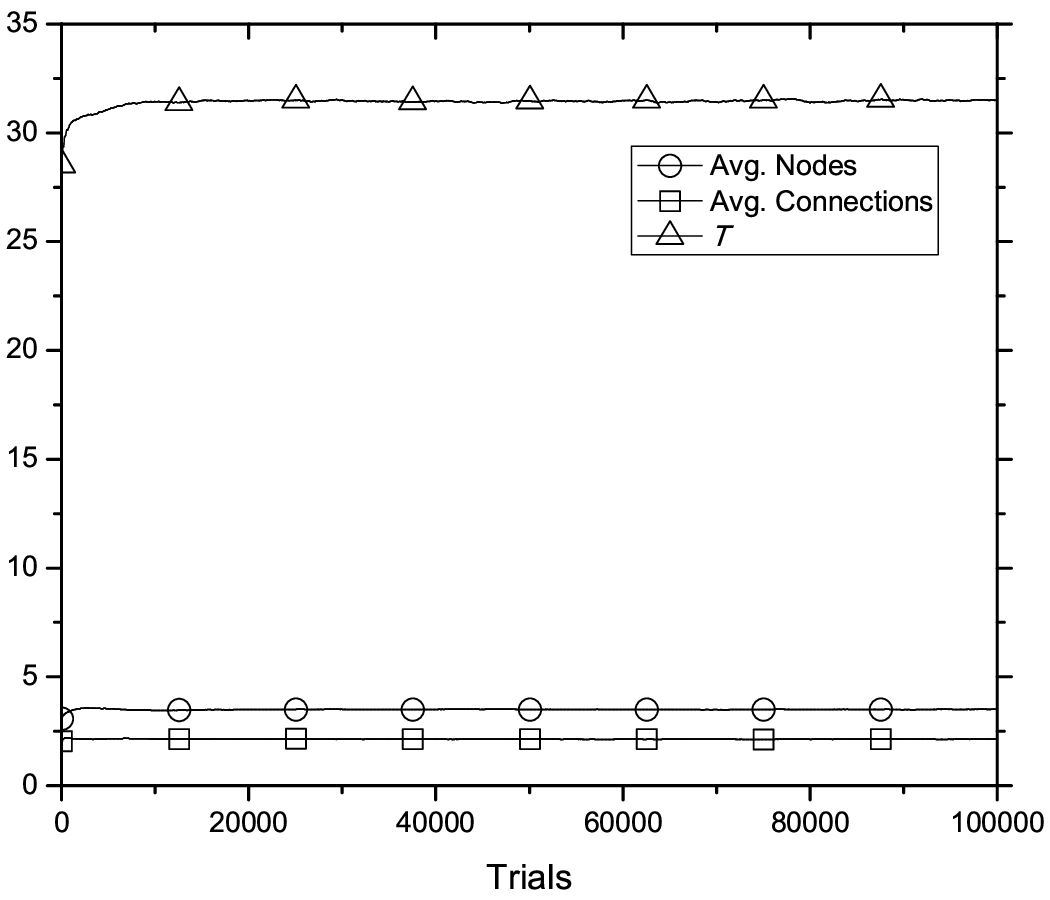,width=\figwidth} }
\caption{fDGP-XCSF Continuous-action frog problem performance.}
\label{fig:fDGP-frog-all}
\end{figure}

Figure~\ref{fig:fDGP-frog-all} illustrates the performance of fDGP-XCSF in the continuous-action frog problem. From Figure~\ref{fig:fDGP-frog-asynch} it can be seen that greater than 99\% performance is achieved in fewer than 4,000 trials ($P=2000$), which is faster than previously reported results \citep[$>$99\% after 30,000 trials, $P=2000$;][]{Tran:2007}, \citep[$>$95\% after 10,000 trials, $P=2000$;][]{Wilson:2007}, and with minimal changes resulting in none of the drawbacks; that is, exploration is here conducted with roulette wheel on prediction instead of deterministically selecting the highest predicting rule, an approach more suitable for online learning. Furthermore, in \cite{Tran:2007} the action weights update component includes the evaluation of the offspring on the last input/payoff before being discarded if the mutant offspring is not more accurate than the parent; therefore additional evaluations are performed which are not reflected in the number of trials reported.

\begin{sloppypar}
From Figure~\ref{fig:fDGP-frog-asynch-top} it can be seen that the average number of (non-unique) macro-classifiers rapidly increases to approximately 1400 after 3,000 trials, before converging to around 150; this is more compact than XCSF with interval conditions \citep[$\sim$1400;][]{Wilson:2007}, showing that fDGP-XCSF can provide strong generalisation. In addition, the networks grow, on average, from 3 nodes to 3.5, and the average connectivity remains static around 1.9. The average mutation rate declines from 50\% to 2\% over the first 15,000 trials before converging to around 1.2\% and the average value of $T$ increases by from 28.5 to 31.5.
\end{sloppypar}

\section{Conclusions}
\label{conclusions}

This paper has explored examples of a temporally dynamic graph-based representation updated with asynchronous parallelism (DGP). The DGP syntax presented consists of each node receiving an arbitrary number of inputs from an unrestricted topology (that is, recursive connections are permitted), and then performing an arbitrary function. The representation is evolved under a self-adaptive and open-ended scheme, allowing the topology to grow to any size to meet the demands of the problem space.

In the discrete case, DGP is equivalent to a form of RBN. It was shown that the XCSF LCS is able to design ensembles of asynchronous RBN whose emergent behaviour can collectively solve discrete-valued computational tasks under a reinforcement learning scheme. In particular, it was shown possible to evolve and retrieve the content-addressable memory existing as locally stable limit points (attractors) within the asynchronously (randomly) updated networks when the final node states from the previous match processing cycle become the starting states for the next environmental input. Furthermore, it was shown that the parameters controlling system sampling of the networks' dynamical behaviour can be made to self-adapt to the temporal complexities of the target environment. The introduced system thus does not need prior knowledge of the dynamics of the solution networks necessary to represent the environment. In particular, the representation scheme was exploited to solve the Woods102 non-Markov maze (that is, without extra mechanisms), a maze which has only previously been solved by LCS using an explicit 8-bit memory register.

A significant advantage of the memory inherent within DGP is that each rule / network's short-term memory is variable-length and adaptive, that is, the networks can adjust the memory parameters, selecting within the limits of the capacity of the memory, what aspects of the input sequence are available for computing predictions. In addition, as the topology is variable length, the maximum size of the short term memory is open-ended, increasing as the number of nodes within the network grows. Thus the maximum size of the content-addressable memory does not need to be predetermined.

Subsequently, the generality of the DGP scheme was further explored by replacing the selectable Boolean functions with fuzzy logical functions, permitting the application to continuous-valued domains. Specifically, the collective emergent behaviour of ensembles of asynchronous FLN were shown to be exploitable in solving continuous-valued input-output reinforcement learning problems, with similar performance to MLP-based neural-XCSF in the continuous-valued multistep grid environment and superior performance to those reported previously in the frog problem.

Current research is exploring the possibilities of DGP as a general representation scheme by which to solve complex problems with LCS. In particular, problems wherein the additional expressiveness of
DGP is most beneficial, such as those requiring memory and temporal prediction.
%
%\bibliographystyle{spbasic}
%\bibliography{references}

\end{document}